\documentclass{article}

\usepackage{microtype}
\usepackage{graphicx}
\usepackage{subcaption}
\usepackage{multirow} 
\usepackage{booktabs} % for professional tables
\usepackage{algorithm}
\usepackage{algorithmic}
\usepackage{hyperref}

\usepackage[accepted]{icml2026}

\usepackage{amsmath}
\usepackage{amssymb}
\usepackage{mathtools}
\usepackage{amsthm}

\usepackage[capitalize,noabbrev]{cleveref}

\theoremstyle{plain}

\theoremstyle{definition}

\theoremstyle{remark}

\usepackage[textsize=tiny]{todonotes}

\icmltitlerunning{Right Predictions, Misleading Explanations}

\begin{document}
\twocolumn[
    \icmltitle{Right Predictions, Misleading Explanations: On the Vulnerability of Vision--Language Model Explanations}

  \begin{icmlauthorlist}
    \icmlauthor{Narges Babadi}{ucalgary}
    \icmlauthor{Hadis Karimipour}{ucalgary}
  \end{icmlauthorlist}

  \icmlaffiliation{ucalgary}{Department of Electrical and Software Engineering, University of Calgary, Calgary, Canada}
  
  \icmlcorrespondingauthor{Narges Babadi}{narges.babadi1@ucalgary.ca}

  \icmlkeywords{Adversarial Machine Learning, Vision--Language Models, CLIP, Explainable AI Security, Explanation Faithfulness, Grey-box Attacks}

  \vskip 0.3in
]
\printAffiliationsAndNotice{}
% this must go after the closing bracket ] following \twocolumn[ ...

% This command actually creates the footnote in the first column listing the
% affiliations and the copyright notice. The command takes one argument, which
% is text to display at the start of the footnote. The \icmlEqualContribution
% command is standard text for equal contribution. Remove it (just {}) if you
% do not need this facility.

% Use ONE of the following lines. DO NOT remove the command.
% If you have no special notice, KEEP empty braces:
%\printAffiliationsAndNotice{}  % no special notice (required even if empty)
% Or, if applicable, use the standard equal contribution text:
% \printAffiliationsAndNotice{\icmlEqualContribution}

\begin{abstract}
Explanation mechanisms are increasingly used to support transparency and trust in vision--language models (VLMs), particularly in settings where model decisions require human oversight. However, the robustness of these explanations remains insufficiently understood. In this work, we investigate whether explanation heatmaps in VLMs, particularly CLIP-based models, faithfully reflect model reasoning under adversarial conditions. We show that explanation maps can be systematically manipulated while preserving the model’s original prediction, revealing a disconnect between predictive behavior and explanation faithfulness. To study this vulnerability, we introduce \emph{X-Shift}, a novel grey-box attack that perturbs patch-level visual representations to redirect explanation heatmaps toward semantically irrelevant regions without altering the predicted output. Unlike conventional adversarial attacks that aim to induce misclassification, X-Shift specifically targets the integrity of the explanation process itself. The attack operates without modifying model parameters and generalizes across multiple CLIP architectures and explanation methods. We evaluate the proposed approach on ImageNet-1k, MS-COCO, and Flickr30K, demonstrating consistent degradation in explanation alignment under imperceptible perturbations while maintaining prediction stability. Furthermore, standard prediction-oriented adversarial attacks fail to reproduce the same explanation-shifting behavior even under substantially larger perturbation budgets. Our findings highlight a fundamental limitation of current explanation mechanisms in VLMs and raise concerns about their use as reliable indicators of model trustworthiness in high-impact applications.
\end{abstract}

\section{Introduction}

Vision--language models (VLMs) have rapidly become foundational components in modern AI systems, enabling joint reasoning over visual and textual modalities. Architectures such as CLIP~\cite{radford2021learning} are now widely deployed across a range of applications, including healthcare, autonomous systems, and decision-support pipelines. In these settings, explanation methods, commonly referred to as explainable AI (XAI), are frequently used to provide transparency by highlighting input regions that influence model predictions. Such explanations are often treated as a proxy for model reliability, supporting tasks such as auditing, debugging, and trust calibration~\cite{lipton2018mythos, li2022exploring, selvaraju2017grad, li2025closer}.

However, a growing body of work has shown that explanation mechanisms can be highly sensitive to adversarial manipulation. Prior studies demonstrate that carefully crafted perturbations can preserve model predictions while substantially altering explanation outputs, thereby breaking the assumed link between explanations and model reasoning~\cite{kindermans2019reliability, ghorbani2019interpretation, dombrowski2019explanations, heo2019fooling, slack2020fooling, lakkaraju2020fool, huang2023focus, ajalloeian2023sparse, kuppa2020black}. These attacks have primarily been studied in unimodal vision models, focusing on gradient-based or post-hoc explanation methods such as LIME and SHAP.

In contrast, the robustness of explanation mechanisms in VLMs remains largely unexplored. This gap is particularly concerning because explanations in VLMs are not merely diagnostic tools; they are often directly integrated into downstream processes such as visual grounding, reasoning, and safety verification. In models such as CLIP, explanations are commonly derived from patch--text similarity maps, which serve as a key signal for interpreting model behavior. If these representations are compromised, both human users and automated systems may be systematically misled without any observable degradation in prediction accuracy. Despite this risk, there are currently limited mechanisms to ensure that such explanations remain faithful under adversarial conditions~\cite{baniecki2024adversarial}.

To illustrate the potential impact, consider a clinical decision-support scenario in which a VLM-based system assists radiologists in diagnosing pneumonia from chest X-rays. The model correctly predicts \emph{pneumonia} and highlights the affected lung region as supporting evidence. Under adversarial manipulation, an attacker can introduce a subtle perturbation that preserves the prediction while redirecting the explanation heatmap toward clinically irrelevant regions. A practitioner relying on the explanation may question the diagnosis or misinterpret the underlying evidence, potentially leading to delayed or incorrect decisions. Notably, such manipulation remains difficult to detect, as standard monitoring mechanisms focus on prediction outputs rather than explanation integrity.

In this paper, we expose a previously underexplored attack surface in VLMs by showing that explanation faithfulness can be manipulated independently of model predictions. We introduced  \textit{X-Shift} , a gray-box adversarial attack that operates on patch-level representations to systematically redirect explanation heatmaps toward attacker-specified regions while preserving the original output. Unlike conventional adversarial attacks that target prediction correctness, X-Shift targets the integrity of explanations, thereby undermining their use as a signal of trust and reliability. The attack requires no access to model parameters or architecture and can be applied at inference time, making it practical in real-world deployment scenarios.

We evaluate X-Shift across multiple datasets and CLIP backbones, demonstrating that it achieves substantial manipulation of explanation maps while maintaining prediction consistency. Furthermore, the attack exhibits strong transferability across architectures and explanation methods, suggesting that the vulnerability is systemic rather than model-specific. Our contributions are summarized as follows:
\begin{itemize}
\item We introduce X-Shift, a gray-box adversarial attack that redirects explanation heatmaps at the patch level while preserving model predictions.
\item We demonstrate that the resulting perturbations remain imperceptible to human observers, enabling stealthy manipulation in practical settings.
\item We show that X-Shift generalizes across multiple VLM architectures and explanation methods, revealing strong transferability and a broad attack surface.
\end{itemize}

\section{Related Work}
\label{sec2}

The vulnerability of deep neural networks to adversarial perturbations is well established~\cite{huang2021exploring, szegedy2013intriguing, goodfellow2014explaining, carlini2017towards, ilyas2018black, ilyas2019adversarial, modas2019sparsefool, babadi2023ensemble, wang2024adversarial, croce2019sparse, madry2017towards}. Most of this literature focuses on degrading predictive performance through misclassification attacks. In contrast, a growing line of work studies a more subtle threat: manipulating model explanations without altering predictions~\cite{baniecki2024adversarial}, challenging the assumption that explanations faithfully reflect model reasoning.

Early studies demonstrated the fragility of explanation methods. Kindermans et al.~\cite{kindermans2019reliability} showed that saliency maps are not invariant to simple input transformations, while Ghorbani et al.~\cite{ghorbani2019interpretation} and Dombrowski et al.~\cite{dombrowski2019explanations} demonstrated that imperceptible perturbations can substantially alter attribution maps without affecting predictions. Beyond input-level attacks, model-level manipulation has also been explored. Heo et al.~\cite{heo2019fooling} trained models to produce misleading attributions under methods such as Grad-CAM and LRP, while Slack et al.~\cite{slack2020fooling} showed that grey-box wrappers can arbitrarily control LIME and SHAP outputs, enabling deceptive practices such as fairwashing~\cite{lakkaraju2020fool}.

Subsequent work has developed more targeted and efficient attacks on explanations. Huang et al.~\cite{huang2023focus} proposed a focus-shifting attack that redirects saliency toward adversary-specified regions while preserving prediction consistency. Ajalloeian et al.~\cite{ajalloeian2023sparse} introduced sparse perturbation techniques that achieve similar effects under minimal input changes. In parallel, Kuppa and Le-Khac~\cite{kuppa2020black} studied grey-box attacks on LIME and SHAP in cybersecurity settings, highlighting the broader risks of explanation manipulation.

Despite these advances, existing work is largely restricted to unimodal image classifiers. The security of explanation mechanisms in vision--language models (VLMs) remains largely underexplored. Prior studies on models such as CLIP have primarily focused on prediction robustness, including adversarial examples, poisoning attacks, and backdoor vulnerabilities~\cite{yang2024clip, zhangmp, huang2025x, jia2022badencoder, jia2021scaling, koh2023grounding, huang2025detecting}. However, these works do not consider the integrity of explanation mechanisms, which are increasingly used as signals for trust and downstream decision-making.

In contrast, we study adversarial manipulation of explanation faithfulness in VLMs. To the best of our knowledge, we provide the first systematic demonstration that CLIP-based explanation maps can be manipulated in a targeted manner while preserving model predictions. Our method operates in a grey-box setting, produces imperceptible perturbations, and generalizes across architectures and explanation methods, revealing a previously unrecognized and practically relevant attack surface in VLM-based systems.

\begin{figure*}[t]
    \centering
    \includegraphics[width=\linewidth]{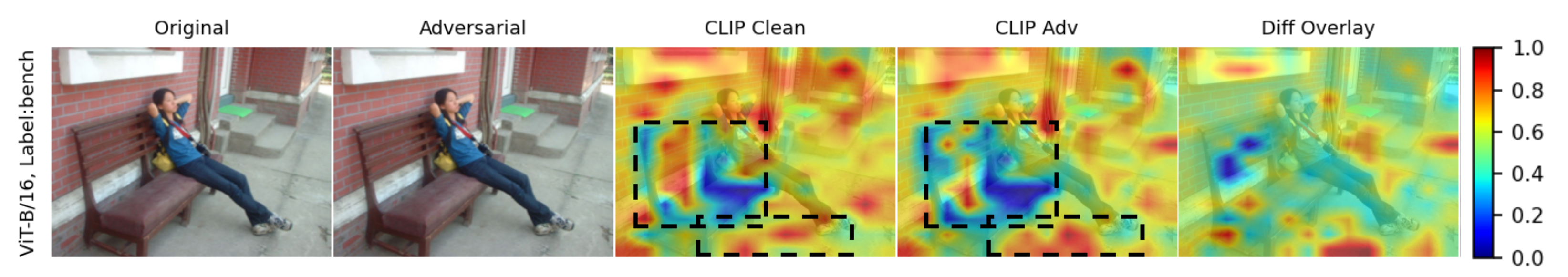}
    \caption{
    Visualization of a sample image under the X-Shift attack.
    From left to right, columns show the clean and adversarial images (the adversarial image is optimized to shift CLIP’s explanation toward ``ground'' while preserving the ``bench'' prediction; see black box), the clean--adversarial difference map, CLIP heatmaps illustrating explanation drift.
    }
    \label{fig:sample}
\end{figure*}

\section{The Proposed Attack: X-Shift}
\label{sec3}

We introduce \textbf{X-Shift}, an explanation-focused adversarial attack that perturbs input images such that model predictions remain unchanged while explanation heatmaps are systematically redirected toward target regions. The attack jointly optimizes four objectives: (i) explanation manipulation, (ii) prediction preservation, (iii) sparsity of perturbations, and (iv) validity constraints on adversarial examples.

\subsection{Threat Model}
X-Shift operates under a \emph{grey-box} threat model. The adversary has access to the model’s forward pass and gradient information sufficient to optimize inputs, but does not modify or access model parameters or training data. This setting is realistic for publicly available vision--language models such as CLIP, where model architectures and weights are accessible, but training data and internal supervision remain unknown. The attack is purely input-driven and does not require retraining or fine-tuning, making it applicable in deployed systems.

\subsection{Baseline: CLIP Model}
\label{secbase}
CLIP \cite{radford2021learning} aligns an image encoder $f_I$ and text encoder $f_T$ in a shared embedding space.  
Given an image $x$ and text $t$, their normalized embeddings are 
$z_I = f_I(x)/\|f_I(x)\|_2,\; z_T = f_T(t)/\|f_T(t)\|_2$, 
with similarity $s(x,t) = z_I^\top z_T$.  
Training minimizes a symmetric contrastive loss over $N$ image–text pairs:
\begin{align}
\mathcal{L}_{\text{CLIP}} = \frac{1}{2N} \sum_{i=1}^N \Bigg[
 -\log \frac{\exp(s(x_i,t_i)/\tau)}{\sum_{j=1}^N \exp(s(x_i,t_j)/\tau)}  \nonumber\\
 -\log \frac{\exp(s(x_i,t_i)/\tau)}{\sum_{j=1}^N \exp(s(x_j,t_i)/\tau)} \Bigg],
\end{align}

where $\tau$ is a learnable temperature.  
Our attack perturbs $x$ into $x_{\text{adv}} = x + \delta$, preserving predictions but shifting explanation maps toward a target class.

\subsection{ X-Shift attack Objectives}
\label{sec:attack_objectives}
We combine the following complementary objectives to achieve explanation-focused adversarial perturbations:

\textbf{Explanation manipulation.}  The primary goal is to force patch embeddings toward the target text embedding. Let $p$ denote the normalized embedding of patch $p$, and $t_{target}$ the target text embedding. Similarity is $s_p = p^\top t_{target}$. We maximize similarity of the top-$K$ patches while suppressing others:
\begin{equation}
\mathcal{L}_{\mathrm{xai}} 
= - \frac{1}{K} \sum_{i \in \text{TopK}} s_{i,t}
  + \alpha \cdot \frac{1}{P-K} \sum_{i \notin \text{TopK}} s_{i,t},
\label{eq:xai_loss}
\end{equation}
where $s_{i,t} = z_i^\top z_{T_{\text{tar}}}$ denotes the similarity between patch embedding $z_i$ and the target text embedding $z_{T_{\text{tar}}}$.

\textbf{Prediction preservation.}  
To prevent label change, we enforce the clean prediction $y^*$ at the global (CLS) level:
\begin{equation}
\mathcal{L}_{\mathrm{pred}} 
= - \log \frac{\exp(z_{\mathrm{cls}}^\top t_{y^*})}
{\sum_{c} \exp(z_{\mathrm{cls}}^\top t_c)}.
\label{eq:pred_loss}
\end{equation}

\textbf{Patch-level margin.} For each patch, the target similarity $s_{p,t}$ must dominate over other classes:
\begin{equation}
\mathcal{L}_{\mathrm{patch}}
= \frac{1}{P} \sum_{p=1}^P 
  \max \big( 0, \max_{c \neq t} (s_{p,c} - s_{p,t} + m) \big),
\label{eq:patch_loss}
\end{equation}
where $s_{p,c} = z_p^\top z_{T_c}$ is the similarity between patch embedding $z_p$ and text embedding $z_{T_c}$.

\textbf{Entropy sharpening.} To avoid diffuse attention maps, we encourage sharp similarity distributions:
\begin{equation}
\mathcal{L}_{\mathrm{entropy}} 
= \sum_{p=1}^P m_p \log m_p, 
\qquad m_p = \frac{\exp(s_{p,t})}{\sum_q \exp(s_{q,t})},
\label{eq:entropy_loss}
\end{equation}
which corresponds to the negative Shannon entropy of the normalized similarities. Minimizing this term encourages sharp and peaked similarity distributions rather than diffuse heatmaps.

\textbf{Sparsity constraint.} Perturbations are restricted to $k$ pixels by projecting $\delta = x_{adv} - x$ onto its top-$k$ entries:
\begin{equation}
\delta \;\leftarrow\; \mathrm{TopK}(\delta, k).
\label{eq:sparsity_constraint}
\end{equation}

\textbf{Validity constraint.}  
Ensure the adversarial image remains in the valid input domain:
\begin{equation}
x_{adv} \in [0,1]^d.
\label{eq:validity_constraint}
\end{equation}

The total objective combines explanation manipulation with auxiliary constraints:  
\begin{equation}
    \label{lss_attack}
    \mathcal{L} 
    = \mathcal{L}_{\mathrm{xai}}
    + \lambda_{\mathrm{pred}} \mathcal{L}_{\mathrm{pred}}
    + \lambda_{\mathrm{patch}} \mathcal{L}_{\mathrm{patch}}
    + \lambda_{\mathrm{ent}} \mathcal{L}_{\mathrm{entropy}}   
\end{equation}

where $\lambda_{\mathrm{pred}}, \lambda_{\mathrm{patch}},$ and $\lambda_{\mathrm{ent}}$ are trade-off coefficients that balance the relative contributions of preserving prediction consistency, enforcing patch-level constraints, and controlling explanation entropy. Tuning these hyperparameters adjusts the strength of each auxiliary objective relative to the main explanation-shifting loss $\mathcal{L}_{\mathrm{xai}}$.

\paragraph{Explainability Attack Algorithm.}
We generate adversarial examples via iterative gradient-based optimization to manipulate explanation maps while preserving the original prediction. The procedure is summarized in \cref{alg:explain_attack_clip}.

\begin{algorithm}[t]
\caption{X-Shift Attack: Explanation Manipulation on CLIP}
\label{alg:explain_attack_clip}
\begin{algorithmic}[1]
\STATE \textbf{Input:} clean image $x$, text embeddings $\{t_c\}$, target index $t$, step size $\eta$, sparsity $k$, iterations $T$
\STATE \textbf{Output:} adversarial image $x^{adv}$
\STATE Initialize $x^{(0)} \gets x$

\FOR{$i = 1$ to $T$}
    \STATE Compute patch embeddings $\{z_p\}$ and CLS embedding $z_{\mathrm{cls}}$
    \STATE Evaluate losses $\mathcal{L}_{\mathrm{xai}}, \mathcal{L}_{\mathrm{pred}}, \mathcal{L}_{\mathrm{patch}}, \mathcal{L}_{\mathrm{entropy}}$
    \STATE Compute total loss:
    \[
    \mathcal{L} \gets \mathcal{L}_{\mathrm{xai}} +
    \lambda_{\mathrm{pred}} \mathcal{L}_{\mathrm{pred}} +
    \lambda_{\mathrm{patch}} \mathcal{L}_{\mathrm{patch}} +
    \lambda_{\mathrm{ent}} \mathcal{L}_{\mathrm{entropy}}
    \]

    \STATE Gradient update:
    \[
    x^{(i)} \gets x^{(i-1)} + \eta \cdot \mathrm{sign}(\nabla_x \mathcal{L})
    \]

    \STATE Sparsity projection:
    \[
    \delta \gets \mathrm{TopK}(x^{(i)} - x^{(0)}, k), \quad
    x^{(i)} \gets x^{(0)} + \delta
    \]

    \STATE Clamp to valid domain:
    \[
    x^{(i)} \gets \mathrm{clip}(x^{(i)}, 0, 1)
    \]
\ENDFOR

\STATE \textbf{return} $x^{adv} = x^{(T)}$
\end{algorithmic}
\end{algorithm}

\section{Experiments}
\label{sec5}

Our experimental evaluation is designed to answer the following research questions:

\begin{itemize}
    \item \textbf{RQ1:} How effectively does X-Shift manipulate patch--text explanation maps while preserving the original model predictions?
    \item \textbf{RQ2:} Does X-Shift maintain prediction consistency across different CLIP backbones and large-scale datasets?
    \item \textbf{RQ3:} To what extent is X-Shift transferable across different CLIP architectures and explanation (heatmap) methods?
\end{itemize}
\begin{figure*}[t]
    \centering
    \includegraphics[width=\linewidth]{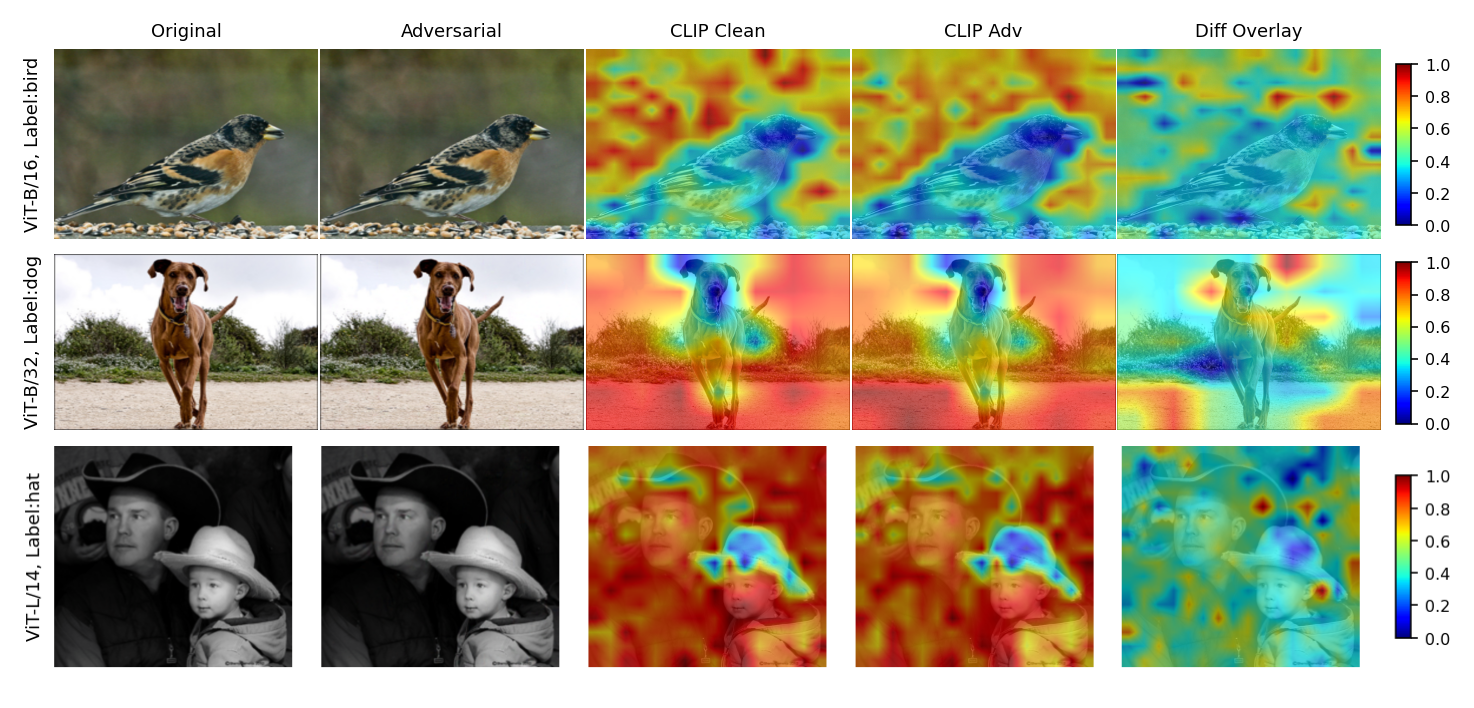}
    \caption{Comparison of CLIP explanations on ImageNet dataset(ViT-B/16, ViT-B/32, ViT-L/14) under X-Shift attack. Columns show original/adversarial images, CLIP heatmaps (clean vs. adversarial).}
    \label{fig:imagenet}
\end{figure*}
\begin{figure*}[t]
    \centering
    \includegraphics[width=\linewidth]{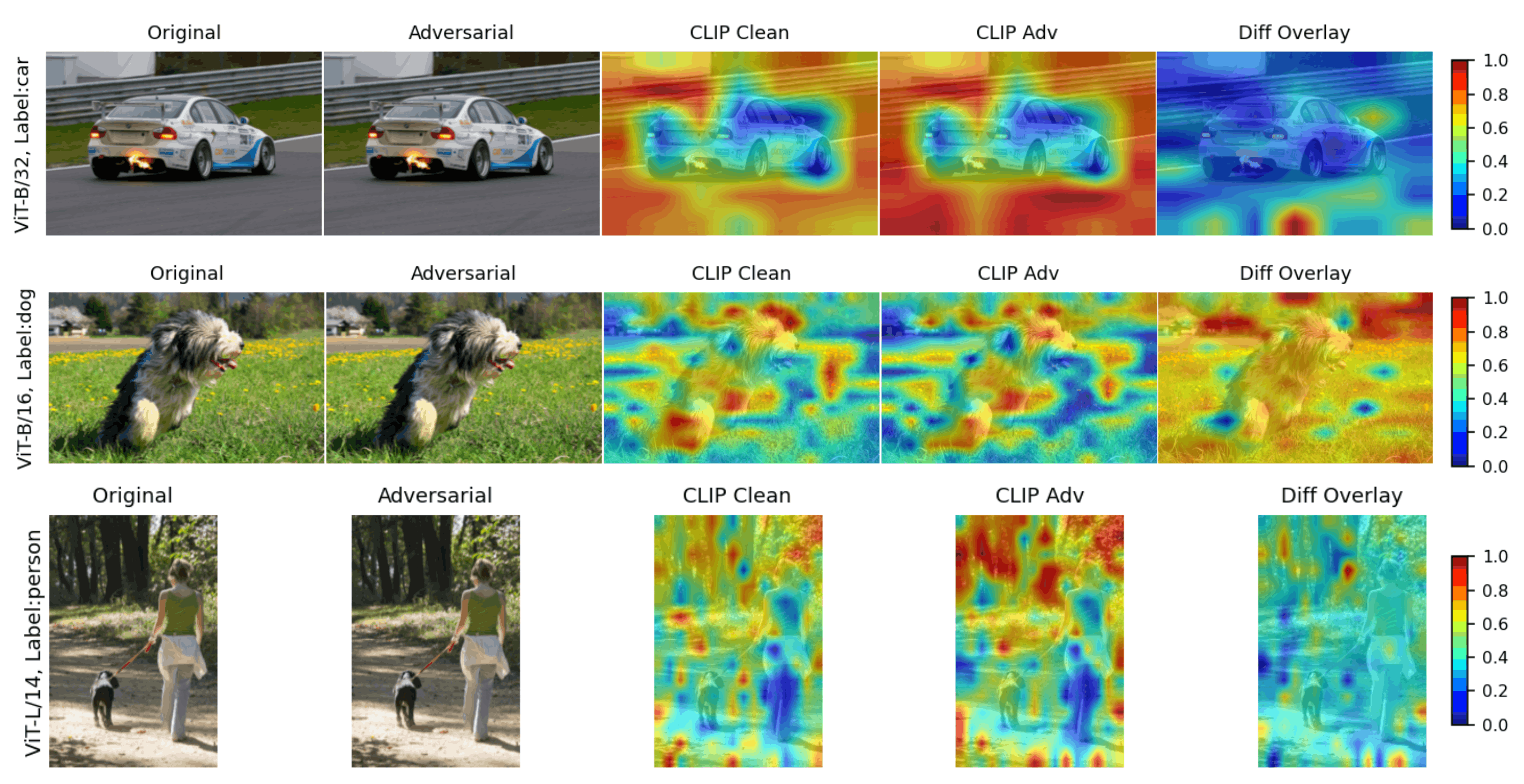}
    \caption{Explanations on Flickr30k samples using CLIP (ViT-B/16, ViT-B/32, ViT-L/14) under X-Shift attack. Shown are original/adversarial images, CLIP heatmaps (clean vs. adversarial).}
    \label{fig:flik}
\end{figure*}

\begin{figure*}[!t]
    \centering
    \includegraphics[width=\textwidth]{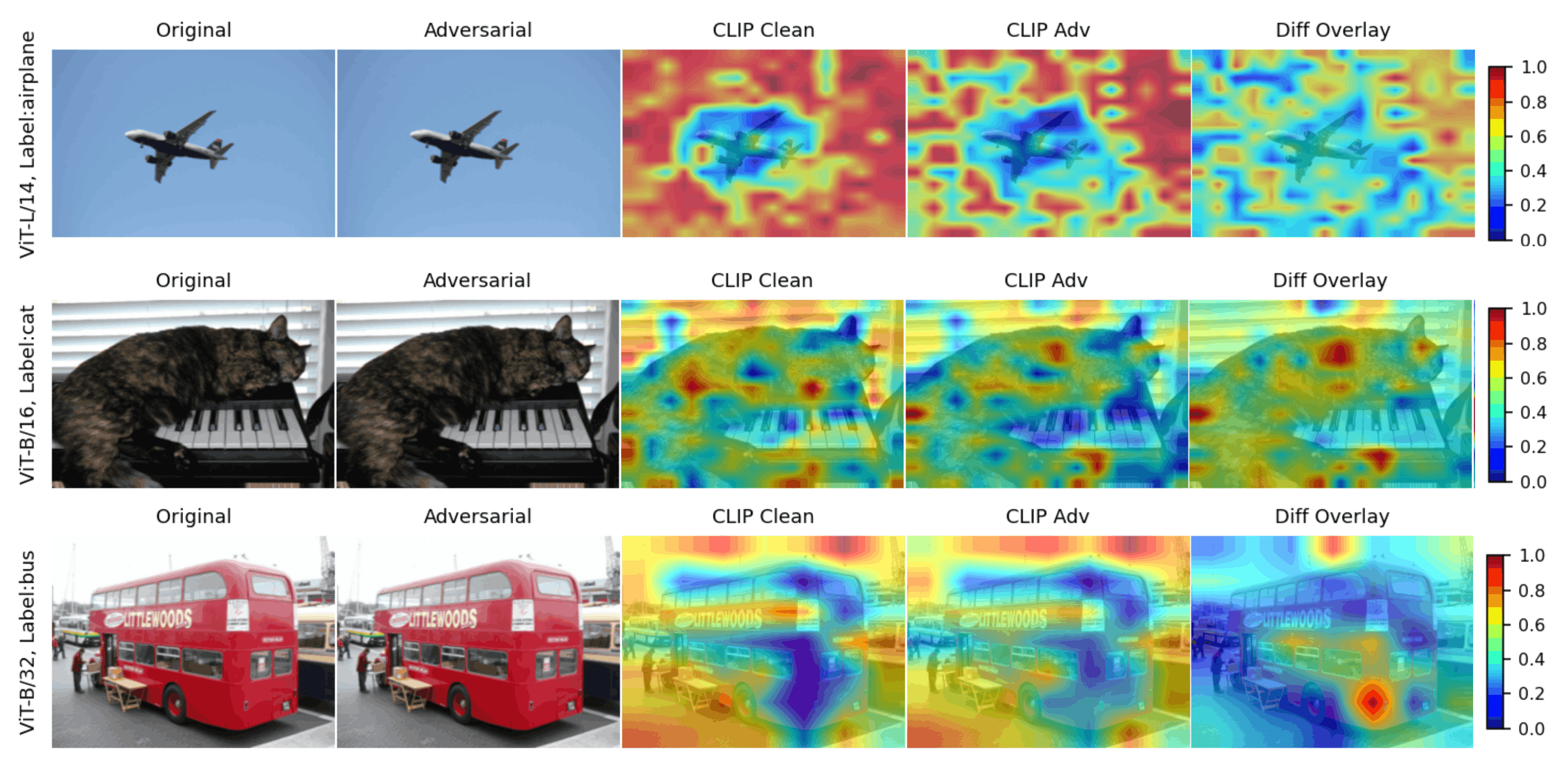}
    \caption{Explanation robustness on COCO samples using CLIP 
    (ViT-B/16, ViT-B/32, ViT-L/14) under X-Shift attack. Columns display original vs.\ adversarial 
    images, CLIP heatmaps (clean vs.\ adversarial).}
    \label{fig:coco_explanations}
\end{figure*}
\textbf{Models and Datasets.} We evaluate our attack at inference time, without requiring additional training data. Experiments are conducted on the validation splits of three benchmark datasets: ImageNet-1k \cite{deng2009imagenet}, Flickr30k \cite{young2014image}, and MS-COCO \cite{chen2015microsoft}, which provide diverse natural images and object-level annotations for assessing VLMs explanations. For models, we utilize the CLIP family of vision–language encoders, specifically ViT-B/16 \cite{radford2021learning}, ViT-B/32 \cite{radford2021learning}, and ViT-L/14 \cite{dosovitskiy2020image}, which span a range of capacities and input resolutions to assess the generality of our attack and defense across different backbones.

\textbf{Implementation.} We implement the proposed attack on official CLIP models, using patch–text similarity maps that compute cosine similarity between patch and text embeddings. Unlike gradient-based attributions (e.g., Grad-CAM, Integrated Gradients), which often yield unstable ViT heatmaps, similarity maps are faithful, text-conditioned, efficient (single forward pass), and deterministic. CLIP employs attention pooling, yielding a $7\times7$ grid for $224\times224$ inputs (datasets resized accordingly). The attack loss follows Section~\ref{sec3}, with weights $20.0$ for $\mathcal{L}_{\mathrm{xai}}$, $\lambda_{\mathrm{ent}}$ for entropy, $\lambda_{\mathrm{margin}}$ for patch separation, and $0.01\lambda_{\mathrm{pred}}$ for prediction consistency, tuned to balance manipulation and stability. 

\textbf{Metrics.} 
We evaluate global prediction stability and explanation robustness using four quantitative metrics:
(i) cosine similarity between clean and adversarial CLS embeddings (CosSim, $\uparrow$), 
(ii) maximum probability deviation (Max $\Delta$Prob, $\downarrow$), and 
(iii) spatial overlap of explanation maps via Top-$k$ Intersection-over-Union (IoU, $\downarrow$). 
Formal definitions are provided in Appendix~\ref{app:metrics}.

\subsection{Results on Explainability}

\textbf{Proposed Attack Effectiveness.} 
Figure~\ref{fig:sample} demonstrates that the X-Shift adversarial perturbations successfully shift CLIP’s explanation maps while preserving the predicted label. In the clean case, the heatmap correctly attends to the input concept (e.g., ``bench''), whereas under the X-Shift attack the attention is redirected toward unrelated regions (e.g., the ``wall''), thereby compromising explanation faithfulness. 

Furthermore, Figures~\ref{fig:imagenet}, \ref{fig:flik}, and \ref{fig:coco_explanations} visualize additional examples from ImageNet, Flickr30k, and COCO. In each case, the perturbation remains imperceptible to humans yet induces substantial shifts in the explanation maps, highlighting the vulnerability of current XAI methods.

Additionally, Table~\ref{tab:comparison-datasets-models} presents a quantitative evaluation of X-Shift under different datasets and CLIP backbones. The results demonstrate that the attack consistently manipulates explanation outputs while preserving model predictions across all settings. Specifically, the cosine similarity values remain high across datasets and architectures, indicating that the global representation and prediction behavior of the model is largely unchanged under attack. At the same time, the Top-$k$ IoU scores exhibit substantial variation, showing that the spatial distribution of explanation heatmaps is significantly altered despite stable predictive outputs. This discrepancy highlights the core effect of X-Shift: decoupling explanation faithfulness from prediction correctness.

\begin{table}[h]

\caption{Quantitative evaluation of \textbf{Vanilla CLIP} under X-Shift attack across datasets and backbones.  
Metrics include cosine similarity (CosSim), maximum probability change under attack (Max $\Delta$Prob), and Top-$k$ IoU.}
\label{tab:comparison-datasets-models}

\centering

\setlength{\tabcolsep}{4pt}

\begin{tabular}{ll|ccc}
\hline
\multirow{2}{*}{\textbf{Dataset}} &
\multirow{2}{*}{\textbf{Backbone}} &
\multicolumn{3}{c}{\textbf{Vanilla CLIP}} \\
\cline{3-5}
& & \textbf{CosSim} & \textbf{Max $\Delta$Prob} & \textbf{IoU} \\
\hline

\multirow{3}{*}{ImageNet}
& ViT-B/16 & 0.805 & 0.004 & 0.487 \\
& ViT-B/32 & 0.807 & 0.004 & 0.450 \\
& ViT-L/14 & 0.948 & 0.000 & 0.551 \\
\hline

\multirow{3}{*}{Flickr30k}
& ViT-B/16 & 0.935 & 0.000 & 0.841 \\
& ViT-B/32 & 0.974 & 0.000 & 0.867 \\
& ViT-L/14 & 0.933 & 0.000 & 0.727 \\
\hline

\multirow{3}{*}{MS-COCO}
& ViT-B/16 & 0.977 & 0.000 & 0.611 \\
& ViT-B/32 & 0.953 & 0.000 & 0.556 \\
& ViT-L/14 & 0.962 & 0.000 & 0.583 \\
\hline

\end{tabular}

\end{table}
Across ImageNet, Flickr30k, and MS-COCO, X-Shift maintains near-identical CosSim and Max $\Delta$Prob values compared to the unperturbed setting, confirming that the attack does not degrade classification confidence or decision consistency. However, the consistent degradation in explanation alignment (as reflected in IoU shifts across backbones) demonstrates that explanation maps can be systematically redirected without triggering changes in model output statistics. Notably, this behavior is preserved across ViT-B/16, ViT-B/32, and ViT-L/14. Overall, these results validate that X-Shift exposes a critical vulnerability in VLMs, where faithful explanations can be adversarially manipulated while maintaining correct predictions, undermining their reliability as a trust signal in safety-critical applications.

\subsection{Transferability of X-Shift Across Vision Transformer Backbones}
\label{sec:transfer-vit}

We evaluate whether explanation-shifting perturbations generated on one CLIP encoder
transfer to other CLIP variants with different patch sizes and embedding dimensions.
Specifically, we test ViT-B/16, ViT-B/32, and ViT-L/14 models in a source-to-target
setting, measuring:

\begin{itemize}
    \item Cosine similarity between clean and adversarial CLS embeddings ($\mathrm{CosSim}_{\text{CLS}}$)
    \item Maximum deviation in predicted probabilities across all text prompts ($\mathrm{Max}\Delta\mathrm{Prob}$)
    \item Patch-level shift in the similarity map for the target concept 
    using $\mathrm{IoU}_{\text{Top-}k}$ (lower is better for measuring explanation manipulation)
    \item Smooth distributional similarity shift using Soft-IoU (also lower is better)
\end{itemize}

\begin{figure}[h]
    \centering
    \includegraphics[width=0.8\linewidth]{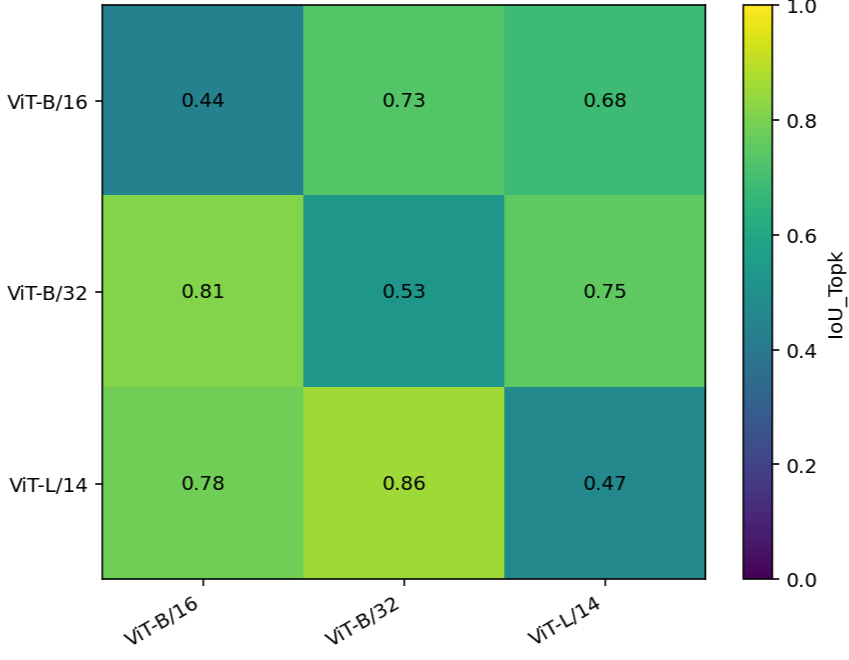}
    \caption{
    Transfer matrix of IoU-TopK (lower indicates stronger manipulation) across
    source $\rightarrow$ target CLIP backbones. ViT-B/32 perturbations transfer
    most broadly, while ViT-L/14 perturbations remain more model-specific.}
    \label{fig:transfer-matrix}
\end{figure}
\begin{table*}[t]
\centering
\scriptsize
\caption{Transferability of explanation-shifting perturbations across CLIP architectures.
We report cosine similarity of CLS tokens (CosSim\,$\uparrow$), maximum change in
predicted probability (Max$\Delta$Prob\,$\downarrow$), and patch-overlap metrics
IoU-TopK and Soft-IoU (both ``lower~is~better'' for capturing successful heatmap
manipulation). Self-attacks achieve the lowest IoU (largest shift), while
cross-model transfer remains moderate but consistent across backbones.
}

\label{tab:transferability}
\begin{tabular}{l l c c c c c c}
\toprule
Source & Target & CosSim$\uparrow$ & Max$\Delta$Prob$\downarrow$ & IoU-TopK$\downarrow$ & Soft-IoU$\downarrow$ & Spearman & EMD \\
\midrule
  & ViT-L/14 & 0.9421 & 0.00044 & \textbf{0.4713} & 0.9837 & 0.7710 & 0.0062 \\
ViT-L/14 & ViT-B/16 & \textbf{0.9928} & \textbf{0.00007} & 0.7818 & 0.9962 & 0.9496 & 0.0010 \\
   & ViT-B/32 & 0.9180 & 0.00023 & 0.8571 & \textbf{0.9973} & \textbf{0.9914} & 0.0017 \\
\midrule
  & ViT-L/14 & 0.9805 & 0.00013 & 0.6842 & 0.9915 & 0.8940 & 0.0024 \\
ViT-B/16 & ViT-B/16 & 0.7628 & 0.00039 & \textbf{0.4412} & 0.9891 & 0.7755 & 0.0104 \\
  & ViT-B/32 & 0.9721 & 0.00029 & 0.7059 & 0.9907 & 0.9194 & 0.0032 \\
\midrule
   & ViT-L/14 & 0.9520 & 0.00017 & 0.6316 & 0.9910 & 0.8743 & 0.0175 \\
ViT-B/32 & ViT-B/16 & 0.9933 & 0.00026 & \textbf{0.5882} & 0.9902 & 0.9283 & 0.0060 \\
   & ViT-B/32 & 0.9278 & \textbf{0.00009} & 0.5750 & 0.9896 & 0.8442 & 0.0128 \\
\bottomrule
\end{tabular}

\end{table*}

\paragraph{Experiment analysis.}
Table~\ref{tab:transferability} shows that self-attacks produce the lowest
IoU-TopK values (0.44--0.47), indicating strong spatial manipulation of the
similarity map without altering model predictions (CosSim\,$>$0.94, Max$\Delta$Prob$<\!4\times10^{-4}$).
Cross-architecture transfer is moderate but consistent: for example,
perturbations crafted on ViT-B/32 transfer to ViT-L/14 with IoU\,=\,0.63,
demonstrating that the attack generalizes across patch sizes (14--32) and
embedding widths. Soft-IoU remains high because CLIP map distributions are smooth,
but localized top-$k$ patch ordering is reliably perturbed. Overall, the results
confirm that X-Shift attacks preserve classification while inducing
model-invariant explanation shifts. Figure~\ref{fig:transfer-matrix} visualizes the mean IoU-TopK transfer matrix
(lower is better), highlighting asymmetric transfer patterns: perturbations from
ViT-B/32 transfer more strongly to other backbones than those from ViT-L/14.
The heatmap corroborates the numerical results in Table~\ref{tab:transferability}
and illustrates which source architectures most reliably induce cross-model
explanation shifts.

%%%%%%%%%%%%%%%%%%%%%%%%%%%%%%%%%%%%%%%%%%%%%%%%%%%%%%%%%%%%%%%%%%%%%%%%%%%%%%%
\subsection{Transferability Across Explainability Methods}
\label{sec:transfer-xai}

\begin{figure*}[!]
    \centering
    \includegraphics[width=\linewidth,height=0.07\textheight]{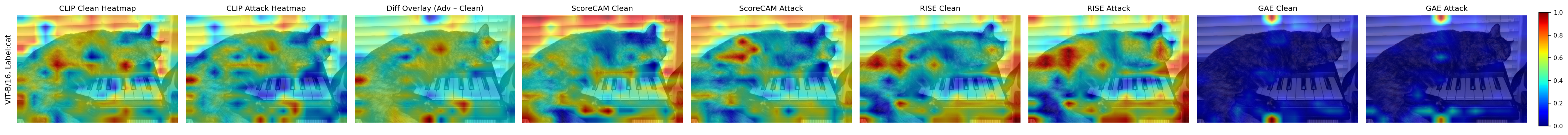}
    \caption{XAI transferability results for the X-Shift adversarial perturbation on COCO dataset with CLIP ViT-B/16. Similarity maps, ScoreCAM, RISE, and GAE all exhibit explanation drift under the attack when applied to vanilla CLIP.}
    \label{fig:tranfer_xai}
\end{figure*}

\begin{figure*}[!]
    \centering
    \includegraphics[width=\linewidth,height=0.07\textheight]{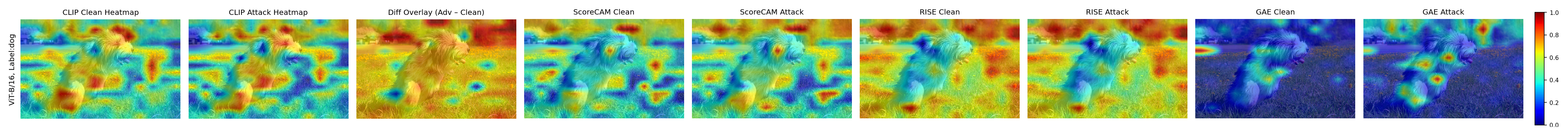}
    \caption{XAI transferability results for the X-Shift adversarial perturbation on the Flickr30k dataset with CLIP ViT-B/16. Similarity maps, ScoreCAM, RISE, and GAE all exhibit explanation drift under the attack when applied to vanilla CLIP.}
    \label{fig:tranfer_xai_flik}
\end{figure*}

\begin{figure*}[!]
    \centering
    \includegraphics[width=\linewidth,height=0.07\textheight]{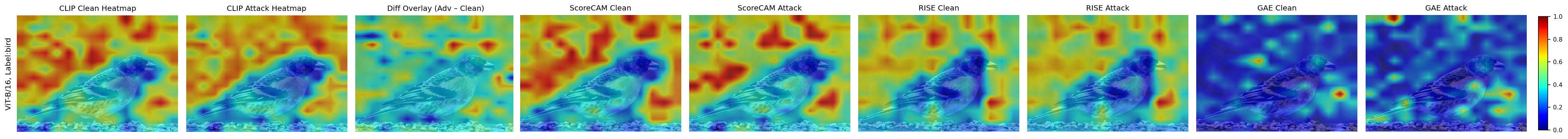}
    \caption{XAI transferability results for the X-Shift adversarial perturbation on the ImageNet dataset with CLIP ViT-B/16. Similarity maps, ScoreCAM, RISE, and GAE all exhibit explanation drift under the attack when applied to vanilla CLIP.}
    \label{fig:tranfer_xai_IMG}
\end{figure*}

To evaluate whether X-Shift attacks generated on vanilla CLIP transfer across attribution methods, datasets, and architectures, we compute a grid of heatmaps using our visualization pipeline. The adversary optimizes the X-Shift perturbation directly on CLIP ViT-B/16, relocating the patch--text similarity mass from the clean concept (e.g., ``cat'') toward an adversarial concept (e.g., ``background'') while preserving the original prediction. Thus, any changes observed in Figures~\ref{fig:tranfer_xai}--\ref{fig:tranfer_xai_IMG} reflect explanation drift rather than classification errors.

\textbf{Choice of XAI methods.} We include ScoreCAM, RISE, and gradient-based explanation (GAE) \emph{solely} to assess attack transferability. These attribution methods were originally designed for single-stream CNN classifiers and do not model CLIP's multimodal text-image alignment or transformer attention. Consequently, they tend to produce diffuse and low-fidelity maps on both clean and adversarial CLIP inputs. Their purpose here is diagnostic: to show that VLMs such as CLIP require dedicated, reliable, and modality-aware explanation tools, and that CNN-based attribution methods lack the grounding needed to produce trustworthy heatmaps for multimodal models.

We evaluate this effect across COCO  (Figure \ref{fig:tranfer_xai}), Flickr30K  (Figure \ref{fig:tranfer_xai_flik}), and ImageNet  (Figure \ref{fig:tranfer_xai_IMG}), plotting similarity maps and their differences for both clean and adversarial images.
In addition to CLIP’s native patch–text similarity heatmaps, we generate ScoreCAM, RISE, and Gradient-based Explanation (GAE) maps to test cross-method transferability.
For vanilla CLIP, the X-Shift perturbation consistently alters the spatial attribution structure: the clean similarity map highlights the true object regions, whereas the adversarial map redirects attention toward background patches aligned with the attacker’s target text.
This drift appears across all datasets and attribution methods, and the difference-overlay visualizations clearly reveal large, structured regions of displaced saliency.

To assess the contribution of each loss component in the X-Shift
objective~(Eq.~\ref{lss_attack}), we conduct a controlled ablation
study reported in full in Appendix~\ref{sec:ablation-loss}.
Briefly, removing $\mathcal{L}_{\mathrm{xai}}$ (\texttt{pred\_only}
variant) yields negligible explanation drift
(IoU$_{\text{Topk}} = 0.7241$, TargetSim $= 0.2364$), confirming that
prediction-stability losses alone cannot drive heatmap manipulation.
Conversely, removing prediction-preservation terms (\texttt{xai\_only})
increases Max$\Delta$Prob by nearly an order of magnitude
($0.000060 \to 0.000540$), producing perturbations detectable by
prediction-monitoring systems and violating the core stealth requirement
of X-Shift. The full objective achieves the best balance: strong
explanation drift (IoU$_{\text{Topk}} = 0.7857$, TargetSim $= 0.2406$)
with a stable CLS representation (CosSim$_{\text{CLS}} = 0.977$) and
negligible classification change (Max$\Delta$Prob $= 0.000066$),
demonstrating that all four loss components are necessary and
non-substitutable for a stable, targeted, and stealthy attack.

\textbf{Comparison with Other Adversarial Attacks.}
To contextualize the behavior of X-Shift, we compare it against standard prediction-targeted adversarial attacks, including FGSM and PGD, under both targeted and untargeted settings. As detailed in Appendix~\ref{app:baseline_comparison}, these baselines are given significantly larger perturbation budgets and stronger optimization settings to maximize their effectiveness. The results show that while conventional attacks succeed in altering model predictions, they do so by substantially distorting the global representation, leading to noticeable degradation in cosine similarity and prediction stability. In contrast, X-Shift operates in a fundamentally different regime: it preserves prediction consistency while achieving substantial manipulation of explanation maps under a much smaller perturbation budget. This comparison highlights that explanation-targeted attacks introduce a distinct and previously overlooked threat model that is not captured by standard adversarial evaluation protocols.
\section{Conclusion}
\label{sec6}

In this work, we investigated the vulnerability of vision--language models, with a particular focus on CLIP, to adversarial manipulation of explanation mechanisms. We proposed \emph{X-Shift}, a targeted grey-box attack that perturbs inputs to systematically distort patch--text similarity heatmaps while preserving model predictions. Our findings demonstrate that explanation outputs can be arbitrarily manipulated without affecting model decisions, exposing a critical disconnect between prediction correctness and explanation faithfulness in current VLMs.

These results highlight the need for faithfulness-aware robustness evaluations and defenses in vision--language systems, particularly in safety-critical applications where explanations are used as proxies for model trustworthiness. Future work will explore the development of defense mechanisms and evaluation frameworks that improve the robustness of explanation methods against such adversarial manipulations, ensuring more reliable deployment of foundation models in real-world settings.

\clearpage
\newpage

\bibliographystyle{icml2026}
\bibliography{example_paper}

\newpage
\appendix
\onecolumn

\section{Evaluation Metrics}
\label{app:metrics}

We measure four complementary aspects of model behavior under X-Shift
perturbations: (i) embedding-level stealth, 
(ii) classifier stability, 
(iii) spatial attribution consistency at the patch level, and 
(iv) distributional similarity of the full explanation map.  
Below we summarize the exact formulations.

\paragraph{Cosine Similarity of CLS Tokens (CosSim\,$\uparrow$).}
This metric quantifies how close the clean and adversarial global embeddings remain.
A high value indicates a \emph{stealthy} attack that preserves high-level semantics.
Given the CLS embeddings $z_{\text{clean}}$ and $z_{\text{adv}}$:

\begin{equation}
    \text{CosSim}_{\text{CLS}} 
    = \frac{z_{\text{clean}} \cdot z_{\text{adv}}}
           {\|z_{\text{clean}}\|_2 \, \|z_{\text{adv}}\|_2}.
\end{equation}

\paragraph{Maximum Probability Deviation (Max$\Delta$Prob\,$\downarrow$).}
This term measures the largest change in predicted probability across all
text prompts.
Low values imply that classification remains unchanged even though the
explanation map shifts:

\begin{equation}
    \text{Max}\,\Delta\text{Prob}
    = \max_{j} 
        \left| 
            P(y_j \mid x_{\text{clean}}) 
            - 
            P(y_j \mid x_{\text{adv}}) 
        \right|.
\end{equation}

\paragraph{Intersection-over-Union of Top-$k$ Patches (IoU-Top$k$\,$\downarrow$).}
We extract the top-$k$ highest-scoring patches in the similarity map for a
target concept, compute the corresponding binary masks
$M_{\text{clean}}$ and $M_{\text{adv}}$, and evaluate. Lower IoU indicates \emph{stronger spatial manipulation}, as fewer top patches
are preserved under the adversarial perturbation.
We use either a fixed $k$ or a percentage $k = \alpha HW$ of all patches:

\begin{equation}
    \text{IoU}_{\text{Top-}k}
    = \frac
        {|M_{\text{clean}} \cap M_{\text{adv}}|}
        {|M_{\text{clean}} \cup M_{\text{adv}}|}.
\end{equation}

\paragraph{Soft Intersection-over-Union (Soft-IoU\,$\downarrow$).}
To capture distributional differences beyond hard top-$k$ sets, we compute a
soft approximation using a temperature $\tau$. This measures global distributional drift, complementing IoU-Top$k$:

\begin{equation}
    p_{\text{clean}} = \text{softmax}(s_{\text{clean}} / \tau), \qquad
    p_{\text{adv}}   = \text{softmax}(s_{\text{adv}}   / \tau),
\end{equation}

\begin{equation}
    \text{Soft-IoU} 
    = \frac{\sum_i \min(p_{\text{clean},i}, p_{\text{adv},i})}
           {\sum_i \max(p_{\text{clean},i}, p_{\text{adv},i})}.
\end{equation}

\paragraph{Spearman Rank Correlation (Spearman).}
We compute the rank correlation between the flattened similarity maps. Low correlation indicates large reordering of influential patches:

\begin{equation}
    \rho = \text{Spearman}(s_{\text{clean}}, s_{\text{adv}}).
\end{equation}

\paragraph{Wasserstein Distance (EMD).}
We compute the Earth Mover’s Distance between flattened similarity scores. EMD captures how much ``work'' is needed to transform the clean explanation
distribution into its adversarial counterpart:

\begin{equation}
    \text{EMD}(s_{\text{clean}}, s_{\text{adv}}) 
    = W_1(s_{\text{clean}},\, s_{\text{adv}}).
\end{equation}

\bigskip

Together, these metrics provide a multi-dimensional characterization of
explanation-shifting behavior: \emph{stealth} (CosSim, Max$\Delta$Prob),
\emph{local spatial reordering} (IoU-Top$k$), and \emph{global distributional drift}
(Soft-IoU, Spearman, EMD).

%%%%%%%%%%%%%%%%%%%%%%%%%%%%%%%%%%%%%%%%%%%%%%%%%%%%%%%%%%%%%%%%%%%%%%%%%%%%%%%
\section{Ablation Study: X-Shift Attack Objective}
\label{sec:ablation-loss}

To validate that each component of the X-Shift objective
(Eq.~\ref{lss_attack}) is necessary, we perform a controlled
ablation over the four loss terms, evaluating three variants:
\texttt{full\_loss} (all terms active), \texttt{xai\_only}
($\mathcal{L}_{\mathrm{xai}}$ alone), and \texttt{pred\_only}
($\mathcal{L}_{\mathrm{pred}} + \mathcal{L}_{\mathrm{margin}}$ only).
Each variant optimizes toward the same target text embedding under
identical hyperparameter settings.

We report four complementary metrics: CLS embedding deviation
(CosSim$_{\text{CLS}}$), maximum class probability change
(Max$\Delta$Prob), top-$k$ patch overlap between clean and adversarial
maps (IoU$_{\text{Topk}}$), and final similarity to the target text
embedding (TargetSim). Together, these metrics isolate each term's
contribution along the three dimensions that define a high-quality
X-Shift attack: \emph{stealthiness}, \emph{prediction stability}, and
\emph{explanation drift}.

\paragraph{Effect of $\mathcal{L}_{\mathrm{xai}}$.}
The \texttt{xai\_only} variant achieves the strongest patch-level drift
(IoU$_{\text{Topk}} = 0.8182$) and highest TargetSim~(0.2405),
confirming that $\mathcal{L}_{\mathrm{xai}}$ is the primary driver of
explanation manipulation. However, removing prediction-preservation
terms increases Max$\Delta$Prob by nearly an order of magnitude
($0.000060 \to 0.000540$ in \texttt{pred\_only}), producing
perturbations that are potentially detectable by prediction-monitoring systems, undermining the core stealth requirement of X-Shift.

\paragraph{Effect of $\mathcal{L}_{\mathrm{pred}}$ and
$\mathcal{L}_{\mathrm{margin}}$.}
The \texttt{pred\_only} variant yields minimal heatmap drift
(lowest IoU$_{\text{Topk}}$) and reduced TargetSim~(0.2364),
demonstrating that prediction-stability losses alone cannot drive
meaningful explanation manipulation. Without $\mathcal{L}_{\mathrm{xai}}$,
patch-text alignment remains largely unchanged, confirming that both
objective families are necessary and non-substitutable.

\begin{figure*}[!t]
    \centering
    \begin{subfigure}{0.32\linewidth}
        \centering
        \includegraphics[width=\linewidth]{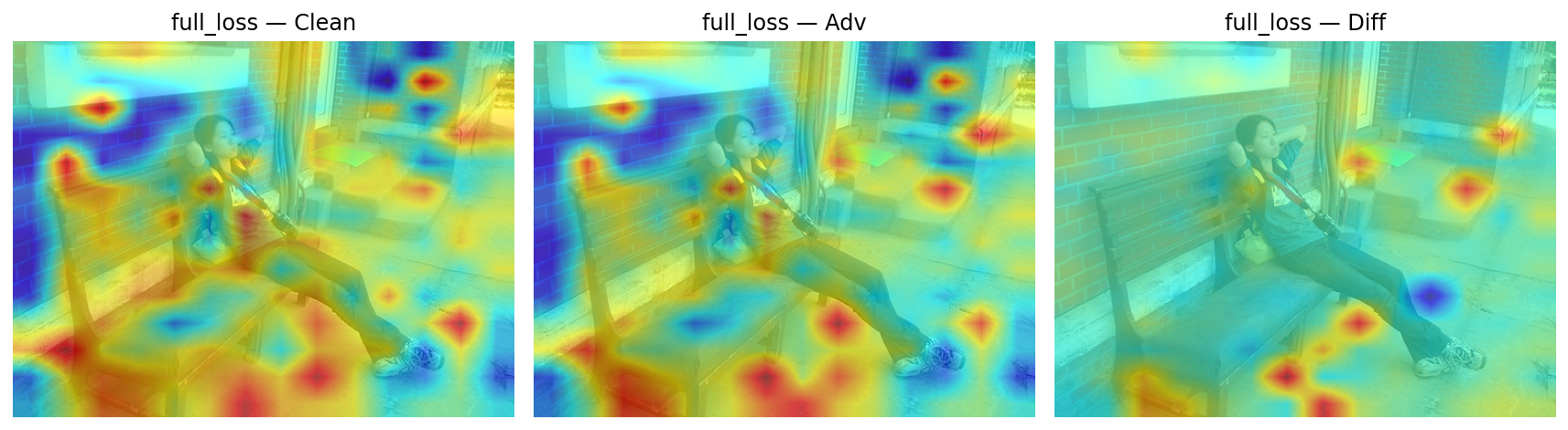}
        \caption{\texttt{full\_loss}: controlled, targeted drift.}
        \label{fig:ablation_full}
    \end{subfigure}
    \hfill
    \begin{subfigure}{0.32\linewidth}
        \centering
        \includegraphics[width=\linewidth]{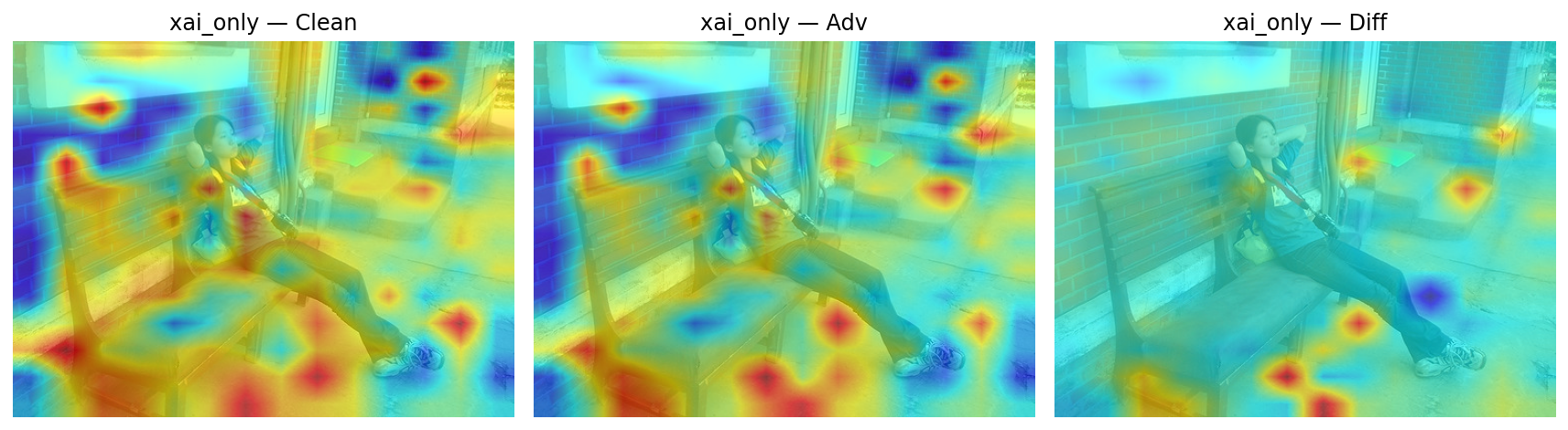}
        \caption{\texttt{xai\_only}: strong but unstable drift.}
        \label{fig:ablation_xai}
    \end{subfigure}
    \hfill
    \begin{subfigure}{0.32\linewidth}
        \centering
        \includegraphics[width=\linewidth]{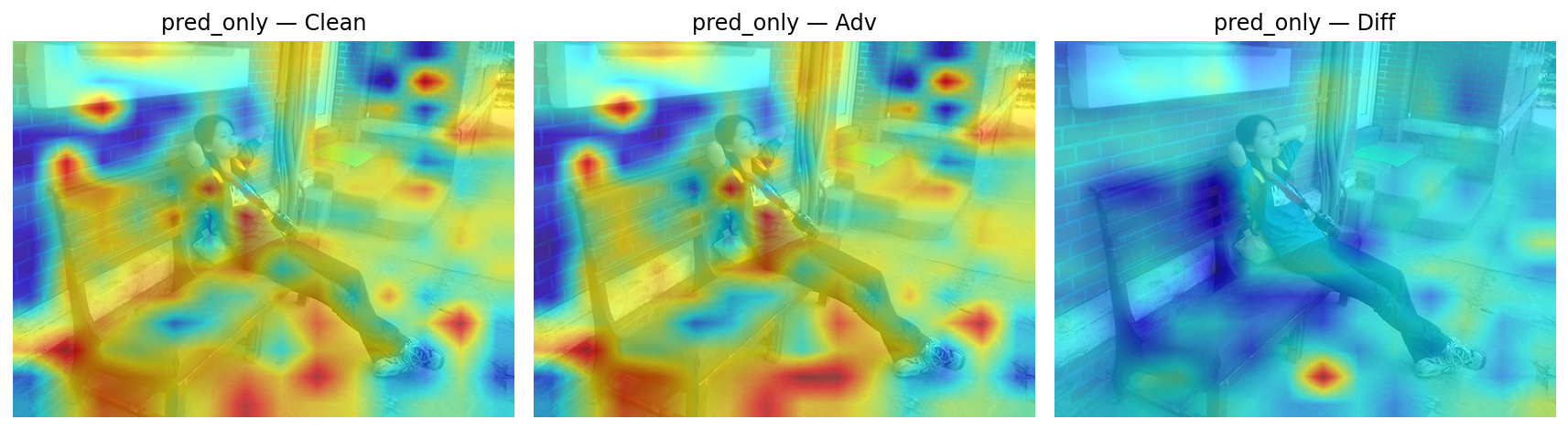}
        \caption{\texttt{pred\_only}: minimal explanation shift.}
        \label{fig:ablation_pred}
    \end{subfigure}
    \caption{Spatial heatmap ablation of X-Shift loss components.
    The full objective produces a balanced, targeted explanation shift
    while preserving the clean prediction. \texttt{xai\_only} yields
    stronger but spatially aggressive drift; \texttt{pred\_only}
    barely alters the explanation map, confirming that both objective
    families are necessary for a stable and stealthy attack.}
    \label{fig:heatmap_ablation}
\end{figure*}

\paragraph{Full Objective.}
The full loss achieves the best balance across all three dimensions:
strong explanation drift, a stable CLS embedding (CosSim$_{\text{CLS}}
= 0.977$), and negligible classification change
(Max$\Delta$Prob $= 0.000066$). Quantitative results are reported in
Table~\ref{tab:ablation}; qualitative heatmap comparisons are shown in
Figure~\ref{fig:heatmap_ablation}.

Spatially, \texttt{xai\_only} produces concentrated but
aggressively redistributed saliency, while \texttt{pred\_only}
preserves most of the clean map structure with minimal drift. The full
objective integrates both behaviors: it yields a controlled yet
significant shift in saliency while maintaining coherent spatial
structure and preserving the original prediction. These results confirm
that all components of Eq.~(\ref{lss_attack}) jointly contribute
to a stable, targeted, and stealthy adversarial perturbation.

\begin{table}[t]
\centering
\scriptsize
\caption{Ablation of X-Shift loss components. The full objective
achieves the best balance of explanation drift, prediction stability,
and stealthiness. $\uparrow$/$\downarrow$ denote preferred direction
for attack quality.}
\label{tab:ablation}
\setlength{\tabcolsep}{4pt}
\begin{tabular}{lcccc}
\toprule
Variant
  & CosSim$_{\text{CLS}}$$\uparrow$
  & Max$\Delta$P$\downarrow$
  & IoU$_{\text{Topk}}$$\downarrow$
  & TargetSim$\uparrow$ \\
\midrule
\texttt{full\_loss}  & 0.977          & 0.000066          & 0.7857 & \textbf{0.2406} \\
\texttt{xai\_only}   & \textbf{0.988} & \textbf{0.000060} & \textbf{0.8182} & 0.2405 \\
\texttt{pred\_only}  & 0.908          & 0.000540          & 0.7241 & 0.2364 \\
\bottomrule
\end{tabular}
\end{table}

\begin{table*}[t]
  \centering
  \caption{%
    Comparison of X-Shift against prediction-targeted adversarial
    baselines on CLIP ViT-L/14 (airplane$\to$ground, COCO) and
    ViT-B/32 (cat$\to$ground, COCO).
    Baselines use $\varepsilon{=}128/255$ and 500 PGD steps;
    X-Shift uses $\varepsilon{=}4/255$ and 400 Adam steps.
    \textbf{Flip\,=\,1}: prediction changed (red);
    \textbf{Flip\,=\,0}: prediction preserved (green).
    X-Shift rows shaded.
    $\uparrow$/$\downarrow$ denote preferred direction.%
  }
  \label{tab:baseline_comparison}
  \setlength{\tabcolsep}{4.5pt}
  \renewcommand{\arraystretch}{1.15}
  
  \begin{tabular}{%
      l                          % Attack
      cccc                       % ViT-L/14 metrics
      cccc                       % ViT-B/32 metrics
    }
    \toprule
    & \multicolumn{4}{c}{\textbf{ViT-L/14} }
    & \multicolumn{4}{c}{\textbf{ViT-B/32}} \\
    \cmidrule(lr){2-5}\cmidrule(lr){6-9}
    \textbf{Attack}
      & IoU$\downarrow$
      & CosSim$\uparrow$
      & Max$\Delta$P$\downarrow$
      & Flip
      & IoU$\downarrow$
      & CosSim$\uparrow$
      & Max$\Delta$P$\downarrow$
      & Flip \\
    \midrule
    FGSM-T ($\varepsilon{=}128/255$)
      & 0.1807 & 0.4732 & 0.0015 & \textcolor{red}{\textbf{1}}
      & 0.3846 & 0.4343 & 0.0014 & \textcolor{red}{\textbf{1}} \\

    FGSM-U ($\varepsilon{=}128/255$)
      & 0.1951 & 0.4447 & 0.0016 & \textcolor{red}{\textbf{1}}
      & 0.0588 & 0.4304 & 0.0014 & \textcolor{red}{\textbf{1}} \\

    PGD-T  ($\varepsilon{=}128/255$)
      & 0.1529 & 0.1853 & 0.0061 & \textcolor{red}{\textbf{1}}
      & 0.1250 & 0.1154 & 0.0041 & \textcolor{red}{\textbf{1}} \\

    PGD-U  ($\varepsilon{=}128/255$)
      & 0.1807 & 0.2908 & 0.0045 & \textcolor{red}{\textbf{1}}
      & 0.0588 & 0.4886 & 0.0029 & \textcolor{red}{\textbf{1}} \\

    \midrule
    \textbf{X-Shift} (ours, $\varepsilon{=}4/255$)
      & \textbf{0.2250} & \textbf{0.8736} & \textbf{0.0003}
      & {\textbf{0}}
      & \textbf{0.5000} & \textbf{0.9549} & \textbf{0.0003}
      & {\textbf{0}} \\
    \bottomrule
  \end{tabular}
\end{table*}

\begin{figure*}[!]
    \centering
    \includegraphics[width=\linewidth]{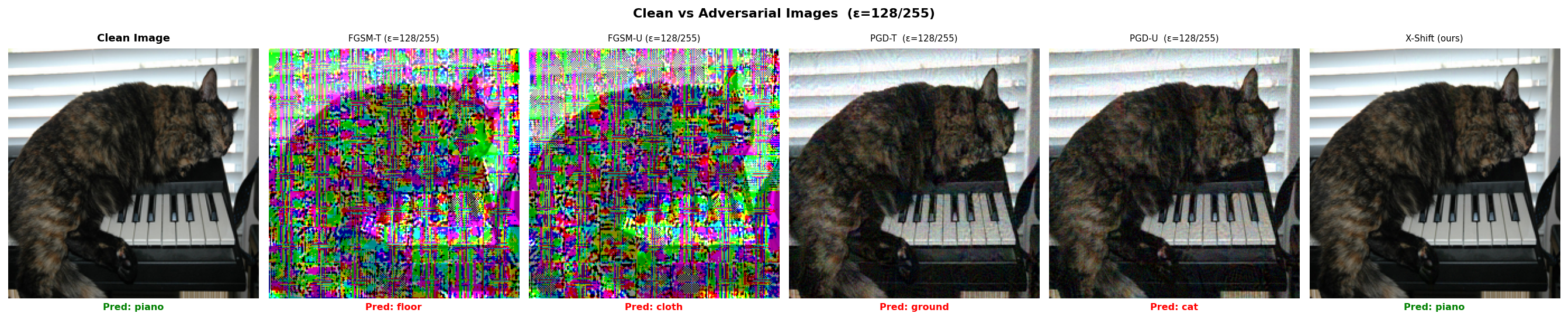}
    \caption{Clean and adversarial examples generated by FGSM and PGD (targeted and untargeted), compared with X-Shift on the COCO dataset using CLIP ViT-B/32.}
    \label{fig:attack_visual_coco}
\end{figure*}

\begin{figure*}[!]
    \centering
    \includegraphics[width=\linewidth]{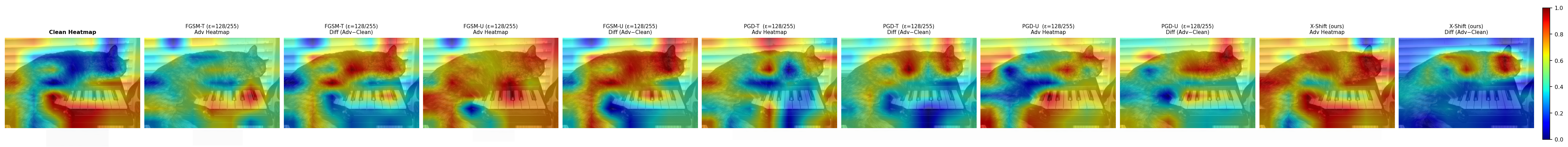}
    \caption{Corresponding explanation (XAI) maps for FGSM, PGD, and X-Shift under targeted and untargeted settings on COCO with CLIP ViT-B/32.}
    \label{fig:attack_xai_coco}
\end{figure*}

\section{Comparison of X-Shift Against Standard Adversarial Attacks}
\label{app:baseline_comparison}

Standard adversarial attacks, such as the Fast Gradient Sign Method (FGSM) \cite{goodfellow2014explaining} and Projected Gradient Descent (PGD), are designed to corrupt model predictions by maximizing classification loss. While these methods are effective in inducing misclassification, they do not explicitly target the model's explanation mechanisms, such as patch-level similarity maps in vision--language models.

In contrast, X-Shift targets a fundamentally different objective: it manipulates explanation heatmaps while explicitly preserving the model's top-1 prediction. This property renders the attack undetectable by standard prediction-monitoring systems. To empirically validate this distinction, we compare X-Shift against four baselines: FGSM targeted (FGSM-T), FGSM untargeted (FGSM-U), PGD targeted (PGD-T), and PGD untargeted (PGD-U), across two CLIP backbones (ViT-L/14 and ViT-B/32).

All methods are evaluated on COCO 2017 images using CLIP's standard preprocessing pipeline. Baselines are given a substantially larger perturbation budget ($\varepsilon = 128/255$) and 500 optimization steps to maximize their ability to flip predictions. In contrast, X-Shift operates under a much smaller budget ($\varepsilon = 4/255$) with 400 optimization steps. We report four metrics: IoU (for explanation overlap), cosine similarity (CosSim) of global representations, maximum probability change (Max$\Delta$P), and a binary Flip indicator (1 if prediction changes, 0 otherwise). By design, X-Shift enforces Flip$=0$.

\paragraph{Results.}
Table~\ref{tab:baseline_comparison} summarizes the results across both backbones. The results reveal a clear separation between prediction-targeted and explanation-targeted attacks. Baseline methods achieve prediction flips (Flip$=1$) but at the cost of significantly degrading the global representation, as reflected in low cosine similarity and higher probability shifts. In contrast, X-Shift preserves prediction consistency (Flip$=0$) while maintaining high cosine similarity and minimal probability change, indicating negligible impact on the model's decision process.

\paragraph{Orthogonal Threat Surfaces.}
These findings demonstrate that prediction-targeted and explanation-targeted attacks operate on fundamentally different threat surfaces. Conventional attacks rely on disrupting the global representation to induce misclassification, which can be detected by monitoring prediction confidence or embedding shifts. In contrast, X-Shift preserves the global representation while selectively manipulating patch-level explanations, bypassing such defenses entirely. This reveals a structural blind spot in current robustness evaluations, which focus exclusively on prediction integrity.

\paragraph{Implications.}
Overall, Table~\ref{tab:baseline_comparison} confirms that X-Shift is not a variant of existing adversarial methods. Also, Figures~\ref{fig:attack_visual_coco} and~\ref{fig:attack_xai_coco} demonstrate that X-Shift is not a variant of existing adversarial methods. It is the only approach that simultaneously (i) achieves targeted manipulation of explanation maps, (ii) preserves model predictions, and (iii) operates under a significantly smaller perturbation budget. These properties define a new class of stealthy attacks that challenge the reliability of explanations as a security and trust signal in vision--language systems.

\end{document}